# Application of Lexical Features Towards Improvement of Filipino Readability Identification of Children's Literature

Joseph Marvin R. Imperial
College of Computer Studies
De La Salle University, National University
Manila, Philippines
joseph_imperial@dlsu.edu.ph

Ethel C. Ong
College of Computer Studies
De La Salle University
Manila, Philippines
ethel.ong@dlsu.edu.ph

## ABSTRACT

Proper identification of grade levels of children's reading materials is an important step towards effective learning. Recent studies in readability assessment for the English domain applied modern approaches in natural language processing (NLP) such as machine learning (ML) techniques to automate the process. There is also a need to extract the correct linguistic features when modeling readability formulas. In the context of the Filipino language, limited work has been done [1, 2], especially in considering the language's lexical complexity as main features. In this paper, we explore the use of lexical features towards improving the development of readability identification of children's books written in Filipino. Results show that combining lexical features (LEX) consisting of type-token ratio, lexical density, lexical variation, foreign word count with traditional features (TRAD) used by previous works such as sentence length, average syllable length, polysyllabic words, word, sentence, and phrase counts increased the performance of readability models by almost a 5% margin (from 42% to 47.2%). Further analysis and ranking of the most important features are shown to identify which features contribute the most in terms of reading complexity.

## 1 Introduction

In the context of language education, the proper assessment of readability levels for young learners is one of the crucial tasks towards learning. This ensures that the reading materials prescribed to them will be at their level of ease when reading, thus, effective comprehension can be achieved while avoiding frustration from the reader [1]. In 2014, the Department of Education (DepEd) launched the Early Grade Reading Assessment (EGRA) program in the Philippines to evaluate the reading competency of kindergarten and grade school students. EGRA is an individually administered oral reading assessment test that aims to assess the literacy capabilities such as letter-sound identification, word decoding, and reading comprehension of young learners in English, Filipino, and other local languages for MTB-MLE subjects. Results of the assessment showed that around 10% to 25% of the surveyed students could not identify correctly letter sounds while 8% to 38% of the students could not read a single word from a given short story [24]. Therefore, research works and efforts on how we can help the nation's reading landscape has never been more important.

Readability identification or readability assessment is the process of mapping reading materials which can be in various forms such as web texts, storybooks, and picture books, to a certain level of ease or difficulty [1]. Usually, this metric is in the form of grade levels where a student belonging to this certain grade should be at ease in reading materials tailored for his/her level. However, this is not a simple task as it is usually done manually and there is a wide range of linguistic features that experts consider in evaluating readability levels of texts [2, 3, 4, 9]. The development of tools that will automatically evaluate and judge the readability of text documents would greatly help linguists, teachers, writers, and researchers in the language education field, and the young learners themselves in acquiring properly labeled reading materials. While the English language is rich in terms of research efforts in automatic readability identification [5, 6, 7, 8], in the context of morphologically-rich languages like Filipino, there are only a few works [1, 2, 21]. Furthermore, these studies still use variations of traditional features for readability formulas.

In this paper, we aim to further the current readability identification formula for Filipino texts by incorporating lexical features such as Type-Token Ratio (TTR), Lexical Density, Lexical Variation such as Noun-Token Ratio and Verb-Token Ratio, and Foreign Words in addition to the traditional features employed in previous studies. We want to discover if these lexical features that are mostly composed of content or information-carrying words that provide meaning to a sentence contribute to the complexity of reading materials in Filipino.

## 2    Previous Work

While research on automatic assessment and readability identification  for the English language is abundant and continuously growing, limited effort is seen on morphologically-rich languages like German, French, Swedish, and Filipino. The earliest readability scale was developed by Flesch [18] in 1948 for English which made use of the average lengths of sentences and words. The score would then be mapped to its equivalent grade level with the use of a conversion table. A newer version of the formula named Flesch-Kincaid Reading Ease was later developed by [8] which no longer gives a numeric value, but instead, the actual suitable grade level itself. The Flesch-Kincaid Reading Ease is one of the most popular readability formulas for English text. It calculates readability based on the average sentence length and average number of syllables in a given text. Likewise, more studies have been conducted using traditional, surface-based features and indicators. Dale and Chall [6] developed a formula based on a long list of 3000 words to determine readability. McLaughlin [7], on the other hand, used a formula based on word count and the number of syllables in sentences. The formula measures readability by the use of polysyllabic (multiple syllable count) word count over a set of sentences. This proved to be useful as words with higher syllable count tend to be present with high level books.

### 2.1    Readability Assessment for Filipino

Up to now, there is still no standardized tool or metric for assessing readability in any local languages in the Philippines. Initial efforts on readability assessment for Filipino were started by Villamin and De Guzman [21] way back in 1979. The formula made use of traditional features such as sentence length and a list containing 3000 words similar to [6] for English. The formula requires a special worksheet for the readability to be identified.

Guevarra [1], on the other hand, made use of linear regression to model seven extracted features including unique word count, total word and sentence count, mean log of word frequency, the average number of syllables, and percentage from a selected top 93 word list. The study produced two readability formulas ($FITRI_{min}$ and $FITRI_{max}$) so that the user can still decide from the range produced by the formula. However, the dataset used in the study was heavily imbalanced, which might have caused bias during training.

In 2014, Macahilig [15] increased the number of features but still used the same linear regression to develop a formula. In the study, other features were highlighted as potential predictors of readability aside from the commonly-used traditional, surface-based features. Lexical features such as part-of-speech, standard spelling system, syllable pattern, prosodic features, and syntactic features were said to greatly influence the readability of Filipino text, and must be explored in future research.

### 2.2    Taking a Step Further with Machine Learning

With the advent of artificial intelligence (AI), new methods of determining readability in texts of various domains have been explored. One of the most used implementations is with the application of machine learning (ML), a subset of AI, for text and document classification. With the help of ML algorithms, recent studies produced state-of-the-art results with traditional features, lexical features, syntactic features, and even up to complex morphological features. Hancke et. al [4] explored the extraction of these said feature sets using a German corpus which obtained high accuracy of 89.7%. A similar favorable result was obtained by Francois and Fairon [9] using a French corpus, 46 selected features, and an SVM approach where the model outperformed previous works that only used traditional features in French as foreign language (FFL) context. In the same manner, Pilan et. al [10] obtained an accuracy at par with the state-of-the-art result in English with a morphologically-rich Swedish dataset. This result was obtained by modeling length-based, lexical, syntactic, semantic, and morphological features using logistic regression, support vector machines, multilayer perceptrons, and decision trees.

## 3    Linguistic Features

For this study, we modelled two different sets of word-based textual features extracted from the reading material corpus: the traditional features (TRAD) and the lexical features (LEX). To the best of our knowledge, no previous work has ever explored the addition of lexical-based features toward the improvement of automatic readability assessment of Filipino texts.

## 3.1 Traditional Features (TRAD)

Traditional or shallow text-based features were extracted as primary features of each reading material. For this study, the extracted TRAD features were the **sentence length** [2, 4, 10, 13, 17], **average token length** [4, 10, 14, 17, 18], **sentence count** and **word count** [1, 8, 6], **average syllables count per word** [1, 4, 8, 14] and **polysyllabic words** [10], for a total of six surface-based features. We noted all previous work that also made use of these features for the development readability indices and formulas in various language domains.

## 3.2 Lexical Features (LEX)

Lexical words are words that belong to a specific lexical category. Lexical categories refer to the classification of lexical items or *'information-carrying'* words according to grammatical criteria such as nouns (*pangngalan*), adjectives (*pang-uri*), verbs (*pandiwa*), and adverbs (*pang-abay*) [19]. The lexical categories are also identified as parts-of-speech (POS) where they can be extracted automatically by a POS Tagger. For this study, we used noun and verb-token ratios which fall under **lexical variation** [4, 14]; **type-token ratio (TTR)** and its variations which are **Root TTR**, **Corrected TTR**, and **Bilogarithmic TTR** [4, 13, 14]; **lexical density** [4, 10, 13, 14]; and the **number of foreign words** for LEX. We noted the previous work that also made use of these features for improvement of readability assessment. We used the Filipino POS Tagger[1] built by Go and Nocon [20] to extract the lexical features mentioned. The formula to calculate each lexical feature is detailed below.

**Lexical Variation (Noun and Verb-Token Ratios)** is the ratio of the total number of words from a lexical category (noun, verb, pronoun) to the total number of words in the text. For this study, we only used two variations, the Noun and Verb-Token Ratio.

$$LV = \frac{Total\ Number\ of\ Specific\ Lexical\ Token}{Total\ Number\ of\ Tokens} \quad (1)$$

**Lexical Density (LD)** is the ratio of the number of words belonging to the four information-carrying lexical categories (nouns, verbs, adverbs, and adjectives) in relation to the total number of words in a text [16]. A high lexical density score signals a high number of information-carrying words while a low lexical density indicates few information-carrying words.

$$LD = \frac{Total\ Number\ of\ Lexical\ Words}{Total\ Number\ of\ Tokens} \quad (2)$$

**Type-Token Ratio (TTR)** measures the number of unique lexical categories or word types $T$ by the total number of words in a text $N$. This sheds light on how a sentence is packed with content-carrying words. A sentence with a TTR score of 1.0 means it is *rich*, where each word is a different lexical category [16]. However, due to its sensitivity of LD and TTR variations to sentence and word count according to [12, 14], variations of TTR were made such as **Root TTR**, **Corrected TTR**, and **Bilogarithmic TTR**. We have extracted these features plus selected only five random sentences for each reading material to be analyzed for uniformity to avoid bias.

$$TTR = \frac{T}{N} \quad (3)$$

$$Root\ TTR = \frac{T}{\sqrt{N}} \quad (4)$$

$$Corr\ TTR = \frac{T}{\sqrt{2N}} \quad (5)$$

$$Bilogarithmic\ TTR = \frac{log(T)}{log(N)} \quad (6)$$

**Foreign Words (FW)** is an additional feature we added which we think may contribute to the complexity of a text. We assume that the number of foreign words (such as English words) present in the text increases as the grade level increases.

$$FW = \frac{Total\ Number\ of\ Foreign\ Words\ (F)}{Total\ Number\ of\ Tokens} \quad (7)$$

We decided to extract these feature sets as one major contribution of this study. The purpose of adding these lexical features is to capture the various aspects of linguistic structures and how it affects the complexity of the text. Our initial assumption is that the more the sentences contain these content-carrying words or high in terms of lexical richness, the higher it is in terms of readability level.

## 4 Experiment Setup

The data used in the study is pre-collected and leveled reading materials for primary education (Levels 1 to 3) from a university library. The dataset is written in modern Filipino, which means that there are some English words present in the text. We rely on some of our selected features to capture this property. Table 1 shows the breakdown of each category along with the total number of tokens, sentences, and reading materials per level. Each material has been evaluated content-wise by an expert on what specific grade level they belong to. The reading materials obtained for each level are in the form of stories. Preprocessing steps such as uniform text type encoding and lowercase handling were performed. We did not yet remove the commas, periods, and punctuation marks as these symbols will serve as delimiters when extracting the traditional, surface-based features.

**Table 1. The distribution of the number of books and token words per class**

| Class | # of Books | # of Tokens | # of Sentences |
|-------|-----------|------------|----------------|
| L1 | 29 | 6,561 | 1,059 |
| L2 | 30 | 13,603 | 1,610 |
| L3 | 30 | 36,022 | 3,330 |
| Total | 89 | 52,186 | 5,999 |

For modelling, we considered the algorithms commonly used in automatic readability assessment by previous work [4, 9, 10, 17, 23] which are the logistic regression and support vector machines. We extracted and modeled traditional and lexical features for each algorithm, testing one feature set at a time and combining both. We also performed two types of feature selection and ranking, the Information Gain ranking and Correlation ranking to highlight which features were highly contributing for identifying readability levels.

**Logistic regression (LR)** makes use of a logistic model (Eq. 8) which represents the probability of a certain class $Pr(Y = c)$ given a set of predictors or features $X$, in this case, the readability level and the traditional and lexical features respectively.

$$Pr\left(Y = c | \vec{X} = x\right) = \frac{e^{\beta_o^{(c)} + x \cdot \beta^{(c)}}}{\sum_c e^{\beta_o^{(c)} + x \cdot \beta^{(c)}}} \qquad (8)$$

**Support vector machines (SVM)** is another algorithm used for classification. SVM constructs a hyperplane in a dimensional space which minimizes the distance between misclassified data points and maximizes the size of the distance to the closest point to the decision boundary. For multiclass approaches such as the readability assessment task, we applied *'one-against-all'* mechanism where we classify the current class against all other classes. The current class or difficulty level $y$ is identified using the argmax formula (Eq. 9) given $x$ as the text features of the current data point and $w_j$ as the weight vectors.

$$y = argmax\ h(x)\ \cdot\ w_j\ +\ \omega_{0,j} \qquad (9)$$

A 10-fold cross-validation was performed to avoid bias. It is also where the evaluation scores of **accuracy**, **macro-averaged F1 score**, and **Root Mean Squared Error (RMSE)** are calculated. The accuracy of the model provides a basis for assessing the performance of the models by giving the ratio of the total number of correctly identified readability levels to the total number of predictions made. F1 score is the weighted averages of recall and precision. We also decided to add RMSE to identify how far the predicted value is to the actual value using the square root of the mean of the prediction errors. In the context of readability identification, previous work such as [1, 2, 13] have used this metric to find the difference between the predicted grade level versus the expected level. The ideal values for accuracy and F1 scores is as close to 1.0 while 0.0 for RMSE.

**Table 2. Classification results using LR**

| Feature Set | # of Features | Classifier Performance | | |
|-------------|---------------|-----------|-----|------|
| | | Accuracy | F1 | RMSE |
| **TRAD** | 6 | 0.377 | 0.365 | 0.489 |
| **LEX** | 8 | 0.330 | 0.315 | 0.657 |
| **TRAD + LEX** | **15** | **0.382** | **0.386** | **0.481** |

**Table 3. Classification Results using SVM**

| Feature Set | # of Features | Classifier Performance | | |
|-------------|---------------|-----------|-----|------|
| | | Accuracy | F1 | RMSE |
| **TRAD** | 6 | 0.420 | 0.413 | 0.503 |
| **LEX** | 8 | 0.290 | 0.290 | 0.546 |
| **TRAD + LEX** | **15** | **0.472** | **0.472** | **0.491** |

## 5 Results and Discussion

We highlight the results of modeling the extracted features for each reading material using the two machine learning algorithms.

### 5.1 Performance of Readability Models

The performances of the various models are reported in Tables 2 and 3. For both the application of logistic regression and support vector machine, we see a similar pattern of increase when we applied the combination of TRAD and LEX features. Although just a small fraction of difference in LR (≈0.01), in SVM, the difference is a significant margin of ≈0.05 or 5%. Similar cases were also seen where SVM outperformed other machine learning algorithms such as LR in previous works for automatic readability assessment for morphologically-rich languages such as German [4], Swedish [10], and French [9]. In addition, SVM is known for handling multiclass classification better than LR using its '*one-against-all*' approach where a current class is treated as a separate class and is classified against the other classes. From the results of the two algorithms, this repetitive binary-like classification subtask turns out to be effective in treating the other classes as one (such as the combination of L2 and L3 versus L1) and then performing the separation using its features.

**Table 4. Confusion matrix of LEX result**

| Actual | L1 (Predicted) | L2 (Predicted) | L3 (Predicted) |
|--------|----------------|----------------|----------------|
| L1 | 9 | 10 | 10 |
| L2 | 12 | 11 | 7 |
| L3 | 16 | 8 | 6 |

**Table 5. Confusion matrix of TRAD result**

| Actual | L1 (Predicted) | L2 (Predicted) | L3 (Predicted) |
|--------|----------------|----------------|----------------|
| L1 | 9 | 11 | 9 |
| L2 | 9 | 14 | 7 |
| L3 | 6 | 10 | 14 |

**Table 6. Confusion matrix of TRAD + LEX result**

| Actual | L1 (Predicted) | L2 (Predicted) | L3 (Predicted) |
|--------|----------------|----------------|----------------|
| L1 | 15 | 7 | 7 |
| L2 | 8 | 12 | 10 |
| L3 | 5 | 10 | 15 |

From Tables 2 and 3, we see the poor evaluation performance of the two models using LEX features alone. This can be cross-referenced with Table 4 and Table 7, showing the confusion matrix of the SVM results and the performance of each model for each grade level respectively. We see in the case of L1 books, two-thirds or 68.9% of its whole dataset has been misclassified to either L2 and L3. The same goes with L2 and L3 books where more misclassifications have occurred than correct classifications, 63.3% misclassification rate for L2 and a high 80% for L3. We deduce that the reason for this is that the lexical features used, such as lexical densities (LD), token ratios (TTRs), and lexical variations (noun and verb token ratios), may only vary by a small degree across all levels on its own; but when we also consider surface or count-based features, there is a significant difference between the readability levels. It is safe to say that as the reader progresses through the levels, the longer the document becomes in terms of sentence count, word count, and phrase counts. This case is observed again in Table 6 where L1 and L3 has the highest correctly classified data count of 15 for both levels, highlighting that there is a significant difference with its TRAD features compared to the other two classes when combined with LEX features.

**Table 7. Table of correct classifications and misclassifications per model per feature**

| Models | L1 | L2 | L3 |
|--------|----|----|----|
| | Corr/Mis | Corr/Mis | Corr/Mis |
| **TRAD** | 31.0% / 68.9% | 46.6% / 53.3% | 46.6% / 53.3% |
| **LEX** | 31.0% / 68.9% | 36.6% / 63.3% | 20.0% / 80.0% |
| **TRAD + LEX** | 51.0% / 49.0% | 40.0% / 60.0% | 50.0% / 50.0% |

Combining TRAD and LEX features as shown in Table 6 slightly increased the number of correctly classified books for each level. This is also observed in Table 7 where L1 and L2 have the highest correctly classified reading materials (51% and 50% respectively) with L2 not far behind. However, going back to Table 6, an exception is seen in L2 and L3 where an equal number of books (10) have been misclassified with each other. We infer that there is significant similarity between L2 and L3 in terms of surface-based text features as well as lexical complexity which caused the misclassification.

## 5.2　Ranking Top Readability Predictors

Taking further the analysis of the results of the models, we performed two types of feature ranking, Information Gain and Pearson Correlation as seen in Table 8. These ranking methods have also been used in previous work [3, 4, 9, 10, 13, 17] to identify which features or feature sets show importance in terms of predicting text complexity. We selected only the Top 10 most contributive features from the TRAD and LEX feature sets.

**Table 8. Top 10 most important features using Information Gain and Correlation**

| Feature Set | Feature | Rank | $\rho$ |
|---|---|---|---|
| TRAD | Polysyllabic Words | 0.148 | 0.115 |
| LEX | TTR | 0.051 | 0.103 |
| | BiTTR | 0.050 | 0.103 |
| | Verb- Token Ratio | 0.022 | 0.094 |
| | CorrTTR | 0.022 | 0.101 |
| | RootTTR | 0.022 | 0.101 |
| TRAD | Word Count | 0.013 | 0.100 |
| | Phrase Count | 0.011 | 0.1 |
| | Sentence Count | 0.006 | 0.1 |
| | Average Syllable Count | 0.002 | 0.1 |

From the table, the number of polysyllabic words or words containing more than 6 syllables scored the highest in terms of information gain and correlation scores. This shows that the higher the syllable count of the word, the more complex it is in terms of readability. For example, L3 books have more polysyllabic words than L2 and L1, and the same relationship for L2 and L1. We solidify this claim in Figure 1 where we identified the frequencies of polysyllabic words for each readability level. The total count of polysyllabic words observes a growth of around 2 to 3 times as the readability level increases. For the mean polysyllabic word count per document, the percentage of growth is still similar.

It can be observed that the addition of the extracted LEX feature set composed of five features, TTR variation and the sole verb-token ratio, were identified as top predictors for readability assessment in Filipino. With this, we infer that there is a significant effect in terms of how much lexical categories (nouns, verbs, conjunctions, adjectives, and pronouns) are present in sentences since it increases in complexity as captured by the TTR scores. The relationship is directly proportional where the higher the difficulty of a sentence in the corpus, the higher the number of lexical categories present in that sentence.

**Figure 1. Frequency count of polysyllabic words**

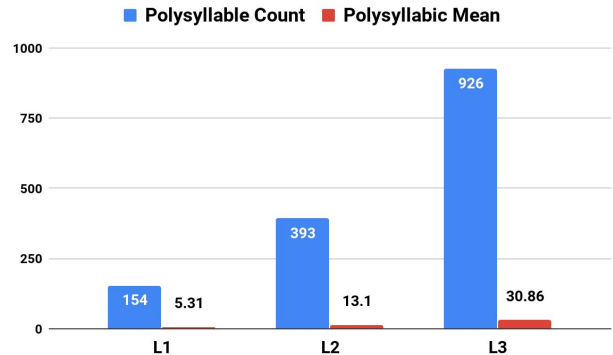

## 6　Conclusion

In this paper, we described a new and improved model for automatic readability assessment of text in Filipino language by combining traditional features (sentence length, average syllable length, polysyllabic words, word, sentence, and phrase counts) with lexical features (type-token ratio, lexical density, lexical variation, foreign word count) and applying the SVM algorithm. Upon analysis of the extracted features, we discovered that lexical features indeed serve as important predictors of text readability in combination with traditional features. The type-token ratios, its variations, and the verb-token ratio were included in the top 10 most informative features of the model. We also note that out of all features, the number of polysyllabic words or words with more than six syllables is the top predictor for readability in primary education reading materials, in which the relationship is directly proportional.

As part of our future work, we intend to extract more linguistic features such as language model perplexities, syntactic features, and the most important one as highlighted by [2], the morphological features to improve the current readability assessment model for Filipino. Filipino language, especially verbs, are known for being complex due to the presence of various affix combinations and duplications in terms of tense, mood, focus, aspect, and voice [22]. In addition, an increase in the number of reading materials which span across several higher levels would definitely help the development of readability assessment models. In terms of content, the same linguistic feature sets can be explored to identify which one serves as a top predictor of readability.